\documentclass{article}
\usepackage{ijcai22}
\usepackage[ijcai,condensebib]{definition}
\usepackage[edges]{forest}

\title{A Survey on Graph Structure Learning: Progress and Opportunities}

\author{
Yanqiao Zhu\textsuperscript{\rm 1,2}\and
Weizhi Xu\textsuperscript{\rm 1,2}\and
Jinghao Zhang\textsuperscript{\rm 1,2}\and
Yuanqi Du\textsuperscript{\rm 3}\\
Jieyu Zhang\textsuperscript{\rm 4}\and
Qiang Liu\textsuperscript{\rm 1,2}\and
Carl Yang\textsuperscript{\rm 5}\And
Shu Wu\textsuperscript{\rm 1,2}\thanks{To whom correspondence should be addressed.}
\affiliations
\textsuperscript{1}Center for Research on Intelligent Perception and Computing\\Institute of Automation, Chinese Academy of Sciences\\
\textsuperscript{2}School of Artificial Intelligence, University of Chinese Academy of Sciences\\
\textsuperscript{3}Department of Computer Science, George Mason University\\
\textsuperscript{4}The Paul G. Allen School of Computer Science and Engineering, University of Washington\\
\textsuperscript{5}Department of Computer Science, Emory University\\
\emails
yanqiao.zhu@cripac.ia.ac.cn\qquad shu.wu@nlpr.ia.ac.cn
}

\begin{document}

\maketitle

\begin{abstract}
Graphs are widely used to describe real-world objects and their interactions.
Graph Neural Networks (GNNs) as a de facto model for analyzing graph-structured data, are highly sensitive to the quality of the given graph structures.
Therefore, noisy or incomplete graphs often lead to unsatisfactory representations and prevent us from fully understanding the mechanism underlying the system.
In pursuit of an optimal graph structure for downstream tasks, recent studies have sparked an effort around the central theme of Graph Structure Learning (GSL), which aims to jointly learn an optimized graph structure and corresponding graph representations.
In the presented survey, we broadly review recent progress in GSL methods.
Specifically, we first formulate a general pipeline of GSL and review state-of-the-art methods classified by the way of modeling graph structures, followed by applications of GSL across domains.
Finally, we point out some issues in current studies and discuss future directions.
\end{abstract}

\section{Introduction}
\label{sec:intro}

Graphs are ubiquitous in representing objects and their complex interactions. As a powerful tool of learning on graph-structured data, Graph Neural Networks (GNNs) have been widely employed for analytical tasks across various domains.
The success of GNNs can be attributed to their ability to simultaneously  exploit the rich information inherent in the graph structure and attributes, however, it is inevitable that the provided graph is incomplete and noisy, which poses a great challenge for applying GNNs to real-world problems.

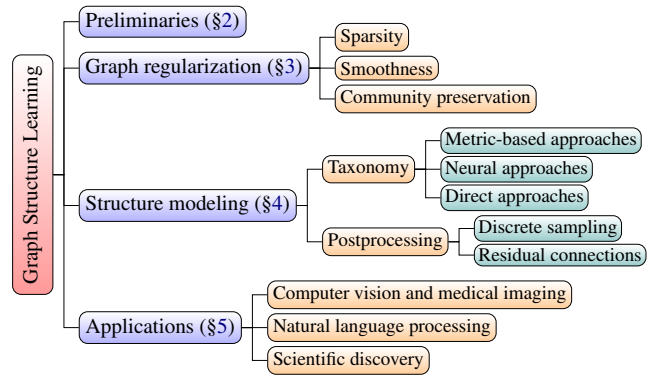
\begin{figure}
\centering
\resizebox{\linewidth}{!}{
\begin{forest}
	for tree={
		draw,
		shape=rectangle,
		rounded corners,
		top color=white,
		grow'=0,
		l sep'=1.2em,
		reversed=true,
		anchor=west,
		child anchor=west,
	},
	forked edges,
	root/.style={
		rotate=90, shading angle=90, bottom color=red!40,
		anchor=north, font=\normalsize, inner sep=0.5em},
	level1/.style={
		bottom color=blue!30, font=\normalsize, inner sep=0.3em,
		s sep=0.2em},
	level2/.style={
		bottom color=orange!40, font=\small, inner sep=0.25em,
		s sep=0.1em},
	level3/.style={
		bottom color=teal!40, font=\small, inner sep=0.2em,
		l sep'=0.5em},
	where n=0{root}{},
	where level=1{level1}{},
	where level=2{level2}{},
	where level=3{level3}{},
	[Graph Structure Learning
		[Preliminaries (\S \ref{sec:preliminaries})
		]
		[Graph regularization (\S \ref{sec:regularization})
			[Sparsity]
			[Smoothness]
			[Community preservation]
		]
		[Structure modeling (\S \ref{sec:model})
			[Taxonomy
				[Metric-based approaches]
				[Neural approaches]
				[Direct approaches]
			]
			[Postprocessing
				[Discrete sampling]
				[Residual connections]
			]
		]
		[Applications (\S \ref{sec:applications})
			[Computer vision and medical imaging]
			[Natural language processing]
			[Scientific discovery]
		]
	]
\end{forest}
}
\label{fig:overview}
\caption{Overview of graph structure learning in this survey.}
\end{figure}

\emph{From the representation learning perspective,}
GNNs compute node embeddings by recursively aggregating information from neighboring nodes \cite{Gilmer:2017tl}; such an iterative mechanism has \emph{cascading effects} --- small noise will be propagated to neighborhoods, deteriorating the quality of representations of many others.
Consider a social network as an example, where nodes correspond to users and edges indicate friendship relations. Fraudulent accounts establish false links to real accounts and thus can easily inject wrong information to the entire network thanks to the recursive aggregation scheme of GNNs, leading to difficulty in estimating account creditability.
In addition, recent research suggests that unnoticeable, deliberate perturbation (aka., adversarial attacks) in graph structure can easily result in wrong predictions for most GNNs \cite{Dai:2018ts,Zhu:2019ik,Zhang:2020tu}.
Thus, highly-quality graph structure is often required for GNNs to learn informative representations \cite{Luo:2021wa}.

\begin{figure*}
	\center
	\includegraphics[width=\linewidth]{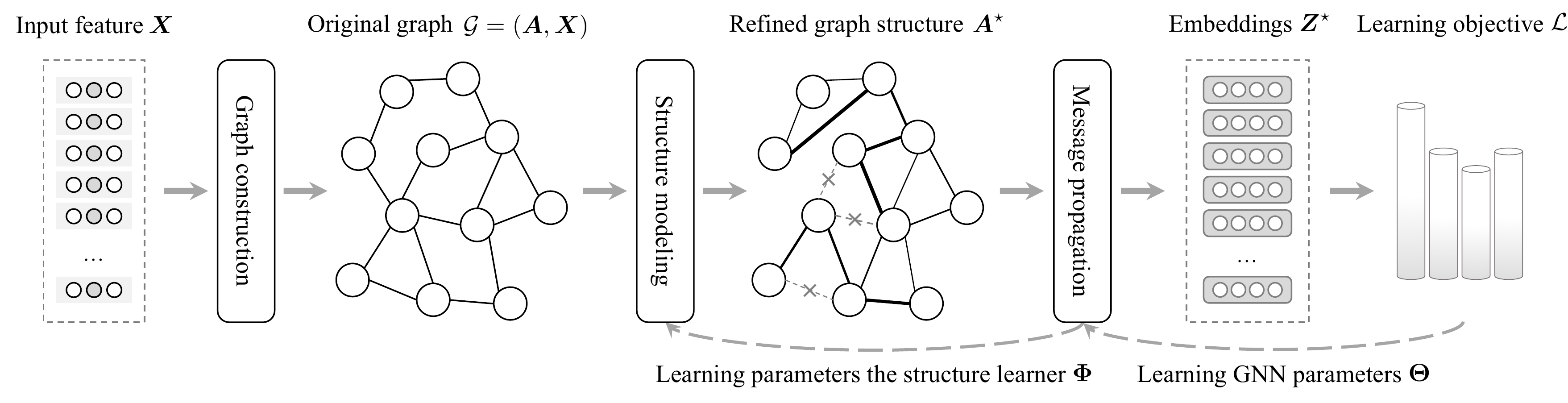}
	\caption{A general pipeline of Graph Structure Learning (GSL). GSL methods start with input features and an optional graph structure. The graph structure is iteratively refined via a structure modeling module. With refined graph structures, graph representations are obtained through Graph Neural Networks (GNNs). Finally, the parameters in GNNs and the structure modeling module are updated alternatively (or jointly) until a preset stopping condition is satisfied.}
	\label{fig:pipeline}
\end{figure*}

\emph{On the application side,}
when modeling the interaction of real-life objects as graphs, we often need to uncover the important substructures (e.g., subgraphs or features) underneath the predictions made by GNNs.
Such model transparency is particularly demanding in safety-critical scenarios such as healthcare.
For example, for explainable brain network analysis, we need to highlight salient regions of interests (nodes in the brain graph) that contribute most to the prediction task \cite{Li:2021fa}.
Besides, with limited prior knowledge, the predefined graph structure only carries partial information of the system, hindering our understanding of the underlying mechanism.
Molecular graph with atoms and bonds as nodes and edges is a notable example. The given graph structures usually overlook non-bonded interactions, which may be of great importance to understand the atom interaction \cite{Leach:2001mm}. 

The aforementioned issues motivate considerable studies around the central theme of \textbf{Graph Structure Learning (GSL)}, which targets at \emph{jointly learning an optimized graph structure and its corresponding representations}.
Unfortunately, many GSL techniques have been proposed across separate research communities and have not yet been systematically reviewed.
In order to establish connections among different research lines and deepen the understanding of the status quo, we present the first survey that has comprehensively reviewed recent progress of GSL.
Specifically, we formulate the problem and present a general pipeline of learning graph structures.
Then, we introduce common graph regularizers, characterize prior studies into three categories, and highlight critical merits of each type.
Furthermore, we introduce applications of GSL in different domains such as computer vision and natural language processing and outline potential research directions for further exploration.

\section{Preliminaries}
\label{sec:preliminaries}

\subsection{Problem Formulation}
\label{sec:formulation}

Let \(\mathcal{G} = (\bm{A}, \bm{X})\) denote a graph, where \(\bm{A} \in \mathbb{R}^{N \times N}\) is the adjacency matrix and \(\bm{X} \in \mathbb{R}^{N \times F}\) is the node feature matrix with the \(i\)-th entry \(\bm{x}_i \in \mathbb{R}^F\) representing the attribute of node \(v_i\).
Given a (partially available) graph \(\mathcal{G}\) at the begining, the goal of GSL is to simultaneously learn an adjacency matrix \(\bm{A}^\star\) and its corresponding graph representations \(\bm{Z}^\star \in \mathbb{R}^{N \times F'}\) that are optimized for certain downstream tasks.

Please kindly note that GSL, although conceptually related, is fundamentally distinct to related problems such as \textbf{graph generation} that target at generating novel, diversified graphs \cite{Du:2021gt}, or \textbf{graph learning} that aims to recover the Laplacian matrix of a homophilous graph corresponding to given node features \cite{Dong:2019gq}.

A typical GSL model involves two trainable components:
(1) a \textbf{GNN encoder} \(f_{\bm{\Theta}}\), receiving one graph as input and producing embeddings for downstream tasks and (2) a \textbf{structure learner} \(g_{\bm{\Phi}}\), modeling the edge connectivity of the graph.
The model parameters \(\{\bm{\Theta}, \bm{\Phi}\}\) are trained with the following learning objective:
\begin{equation}
	\mathcal{L} = \mathcal{L}_\text{tsk}(\bm{Z}^\star,\bm{Y}) + \lambda \mathcal{L}_\text{reg}(\bm{A}^\star, \bm{Z}^\star, \mathcal{G}),
	\label{eq:objective}
\end{equation}
where the first term refers to a task-specific objective with respect to the ground truth \(\bm{Y}\), the second term imposes prior constraints on the learned graph and its representations, and \(\lambda\) is a hyperparameter. %

\subsection{Graph Structure Learning Pipeline}
As shown in Figure \ref{fig:pipeline}, most existing GSL models follow a three-stage pipeline: (1) graph construction, (2) graph structure modeling, and (3) message propagation.

\paragraph{Graph construction.}
Initially, when the given graph structure is incomplete or even unavailable at all, we construct a preliminary graph as a starting point.
There are a variety of ways to build such initial graph, among which the two most common options are (1) \(k\) Nearest Neighbors (\(k\)NN graphs) \cite{Preparata:1985cg} and (2) \(\epsilon\) proximity thresholding (\(\epsilon\)-graphs) \cite{Bentley:1977kt}.
Both approaches compute pairwise distance of node features using kernel functions at first.
For the \(k\)NN graphs, we connect two nodes \(v_i, v_j\) if \(v_i\) is among the \(k\)-closest neighbors of \(v_j\).
For the latter \(\epsilon\)-graphs, we create an edge between two nodes if their similarity is smaller than a preset threshold \(\epsilon\). 

\paragraph{Graph structure modeling.}
The core of GSL is the structure learner \(g\) that models the edge connectivity to refine the preliminary graph.
In this work, we categorize existing studies into the following three groups:
\begin{itemize}
	\item \textbf{Metric-based approaches} employ a metric function on pairwise node embeddings to derive edge weights.
	\item \textbf{Neural approaches} leverage more expressive neural networks to infer edge weights given node representations.
	\item \textbf{Direct approaches} treat the adjacency matrix as free learnable parameters and directly optimize them along with GNN parameters \(\bm\Theta\). 
\end{itemize}
Unlike direct approaches, metric-based and neural approaches learn edge connectivities through \emph{a parameterized network}, which receives node representations and produces a matrix \(\widetilde{\bm{A}}\) representing the optimized graph structure.
After obtaining a primitive graph structure represented via the structure learners, extra \textbf{postprocessing steps} such as discrete sampling may be involved to produce the final graph \(\bm{A}^\star\).

\paragraph{Message propagation.}
After obtaining an optimized adjacency matrix \(\bm{A}^\star\), we employ a GNN encoder \(f\) to propagate node features to the refined neighborhoods, which results in embeddings \(\bm{Z}^\star\).

Notably, it is common to repeat the last two steps until specified criteria are satisfied due to the difficulty of optimizing the graph structure.
In other words, the representations resulting from message propagation will be utilized to model edge weights in the second step, iteratively refining both graph topologies and representations.

\section{Graph Regularization}
\label{sec:regularization}

In practice, we often want the learned graphs to satisfy certain properties.
Prior to graph structure modeling techniques, we review three general graph regularization techniques in this section: sparsity, smoothness, and community preservation.

\subsection{Sparsity}
\label{sec:sparsity}

Motivated by the observation that real-world graphs have noisy, task-irrelevant edges, we usually enforce a sparsity constraint on the adjacency matrix.
A common idea is to penalize the \(\ell_0\)-norm, which corresponds to the number of nonzero elements:
\begin{equation}
	\mathcal{L}_\text{sp}(\bm{A}) = \|\bm{A}\|_0.
	\label{eq:sparsity}
\end{equation}
However, minimizing Eq. (\ref{eq:sparsity}) is generally NP-hard. Therefore, we usually resort to \(\ell_1\)-norm, its convex relaxation, which can be solved using coordinate descent \cite{Wu:2008cd} or proximal gradient descent \cite{Beck:2010ga}.
Another way to incorporate the sparsity constraint is by its continuous relaxation.
\citet{Louizos:2018tl} propose a differentiable estimator for \(\ell_0\)-norm via variational optimization and concrete relaxation.

Besides directly panelizing the non-zero entries, some resort to implicit sparsification as well, e.g., hard thresholding the adjacency matrix similar to constructing \(\epsilon\)-graphs.
Discrete sampling as mentioned in \S\ref{sec:postprocessing} is also helpful for ensuring sparsity of the learned graphs.

\subsection{Smoothness}
\label{sec:smoothness}

Consider each of the \(N\) columns of the node feature \(\bm{X}\) as a signal on the graph \(\mathcal{G}\).
A widely used assumption for graph signal processing is that signals change smoothly between adjacent nodes \cite{Ortega:2018ea}.
To enforce smoothness of the graph signals, we minimize the following term:
\begin{equation}
	\mathcal{L}_\text{sm}(\bm{X}, \bm{A}) = \frac{1}{2} \sum_{i,j=1}^N \bm{A}_{ij} (\bm{x}_i - \bm{x}_j)^2 = \tr(\bm{X}^\top \bm{L} \bm{X}),
	\label{eq:smoothness}
\end{equation}
where \(\bm{L} = \bm{D} - \bm{A}\) and \(\bm{D}\) is the degree matrix of \(\bm{A}\).
An alternative form of Eq. (\ref{eq:smoothness}) is to use the normalized graph Laplacian \(\widehat{\bm{L}} = \bm{D}^{-\frac{1}{2}}\bm{LD}^{-\frac{1}{2}}\), which makes the feature smoothness independent of node degrees \cite{Chung:1997sp}.
Theoretically, minimizing Eq. (\ref{eq:smoothness}) penalizes edges connecting distant rows of \(\bm{X}\), indicating that graphs corresponding to smooth signals have a sparse edge set.
Therefore, we may impose an auxiliary connectivity constraint \cite{Kalofolias:2016tf}:
\begin{equation}
	\mathcal{L}_\text{con}(\bm{A}) = - \bm{1}^\top \log(\bm{A1}),
	\label{eq:connectivity}
\end{equation}
where the logarithmic barrier forces the degrees to be positive, but does not panelize individual edges to be sparse.
Though minimizing Eq. (\ref{eq:connectivity}) leads to sparse graphs, changing its coefficient does not control the degree of sparsity but only scales the solution.
To directly control the sparsity, it is suggested to use sparsity regularizers (\S \ref{sec:sparsity}) together with  \(\mathcal{L}_\text{con}\).

\subsection{Community Preservation}
\label{sec:community}

Intuitively, for real-life graphs nodes in different topological clusters tend to have different labels. As a result, edges spanning multiple communities could be regarded as noise \cite{Zhu:2020ui}.
From graph spectral theory we know that the rank of one adjacency matrix relates to the number of connected components in that graph and low-rank graphs have a densely connected component \cite{Biggs:1993ag}.
Therefore, to remove potentially noisy edges and best preserve community structures, we introduce a low-rank regularization
\begin{equation}
	\mathcal{L}_\text{cp}(\bm{A}) = \rank(\bm{A}).
	\label{eq:rank}
\end{equation}
Due to the discrete nature of the rank operator, matrix rank minimization is hard. One often seeks to optimize the nuclear norm in place of Eq. (\ref{eq:rank}), which is defined as the sum of its singular values.
This heuristic has been observed to produce low-rank solutions in practice \cite{Fazel:2001rm} and can be optimized efficiently (e.g., via singular value thresholding \cite{Cai:2010hz}).

\begin{table*}[t]
	\centering
	\small
	\caption{Summary of representative graph structure learning methods.}
	\begin{adjustbox}{max width=\textwidth}
	\begin{threeparttable}
	\def\tnote#1{\textsuperscript{[\TPTtagStyle{#1}]}}%
	\label{tab:model}
	\begin{tabular}{cccccccccccc}
	\toprule
	& \multirow{2.5}[0]{*}{Method} & \multirow{2.5}[0]{*}{Reference} & \multirow{2.5}[0]{*}{Graph construction} & \multirow{2.5}[0]{*}{Structure modeling} & \multicolumn{2}{c}{Postprocessing} & \multicolumn{4}{c}{Graph regularization} & \multirow{2.5}[0]{*}{Training} \\
	\cmidrule(lr){6-7} \cmidrule(lr){8-11}
	&  &       &       &       & Sampling & Residual & SP\tnotex{fn:sparsity} & SM\tnotex{fn:smoothness} & CON\tnotex{fn:connectivity} & CP\tnotex{fn:community} &  \\
	\midrule
	\parbox[t]{0mm}{\multirow{6}{*}{\rotatebox[origin=c]{90}{Metric-based}}}
	& AGCN & \cite{Li:2018wu}  & ---   & Gaussian kernel &       & \cmark &       &       &       &       & {\color{green}\(\bullet\)} End-to-end \\
	& GRCN & \cite{Yu:2020vw} & ---   & Inner product &       & \cmark & \cmark &       &       &       & {\color{green}\(\bullet\)} End-to-end \\
	& IDGL & \cite{Chen:2020wu}  & ---   & Cosine similarity &       & \cmark & \cmark & \cmark & \cmark &       & {\color{green}\(\bullet\)} End-to-end \\
	& HGSL & \cite{Zhao:2021vn} & \(\epsilon\)-graphs & Cosine similarity & & \cmark & \cmark & & & & {\color{green}\(\bullet\)} End-to-end \\
	& GDC & \cite{Klicpera:2019vc}  & ---   & Diffusion kernel &       &       & \cmark &       &       &       & {\color{green}\(\bullet\)} End-to-end \\
	& AM-GCN & \cite{Wang:2020bs} & \(k\)NN graphs & Multiple kernels &       &       &       &       &       &       & {\color{green}\(\bullet\)} End-to-end \\
	\midrule
	
	\parbox[t]{0mm}{\multirow{4}{*}{\rotatebox[origin=c]{90}{Neural}}}
	& GLCN  & \cite{Jiang:2019wp} & \(k\)NN graphs & One-layer neural net &       &       & \cmark & \cmark &       &       & {\color{green}\(\bullet\)} End-to-end \\
	& PTDNet & \cite{Luo:2021wa} & ---   & Multilayer perceptron & \cmark &       & \cmark &       &       & \cmark & {\color{green}\(\bullet\)} End-to-end \\
	& VIB-GSL & \cite{Sun:2022wx} & ---   & Dot-product attention & \cmark &       & \cmark &       &       &       & {\color{green}\(\bullet\)} End-to-end \\
	& SAN   & \cite{Kreuzer:2021wi} & ---   & Transformer + Laplace PE &       &       &       &       &       &       & {\color{green}\(\bullet\)} End-to-end \\
	\midrule

	\parbox[t]{0mm}{\multirow{3}{*}{\rotatebox[origin=c]{90}{Direct}}}
	& GLNN  & \cite{Gao:2020em} & ---   & Free variables &       &       & \cmark & \cmark &       &       & {\color{green}\(\bullet\)} End-to-end \\
	& ProGNN & \cite{Jin:2020br} & ---   & Free variables &       &       & \cmark & \cmark &       & \cmark & {\color{orange}\(\bullet\)} Alternative \\
	& LRGNN & \cite{Xu:2021fs} & ---   & Free variables &       &       &       & \cmark &       & \cmark & {\color{orange}\(\bullet\)} Alternative \\

	\bottomrule
	\end{tabular}
	\begin{tablenotes}[flushleft,para]
	\footnotesize{
		Graph regularization:
		\item[$\star$]\label{fn:sparsity} Sparsity (SP, \S\ref{sec:sparsity})
		\item[\dag]\label{fn:smoothness} Smoothness (SM, \S\ref{sec:smoothness})
		\item[\ddag]\label{fn:connectivity} Connectivity (CON, \S\ref{sec:smoothness})
		\item[$\sharp$]\label{fn:community} Community preservation (CP, \S\ref{sec:community})
	}
	\end{tablenotes}
	\end{threeparttable}
	\end{adjustbox}
\end{table*}

\section{Graph Structure Modeling}
\label{sec:model}

In this section, we examine representative GSL models and discuss their common structure modeling techniques.
At first, we succinctly discuss three categories of graph structure modeling, which obtains an intermediate graph optimized for certain end tasks.
After that, we introduce techniques for postprocessing that learned graph structure.
For each category, we summarize representative models in Table \ref{tab:model}.

\subsection{Model Taxonomy}

\subsubsection{Metric-based Approaches}

Metric-based approaches use kernel functions to compute the similarity between node feature/embedding pairs as the edge weights.
According to the network homophily assumption that edges tend to connect similar nodes \cite{Newman:2018ne}, these approaches refine the graph structure by promoting intra-class connections, leading to more compact representations.
Owing to the differentiability of kernel functions, most metric-based approaches enjoy the capability of end-to-end training.

\paragraph{Gaussian kernels.}
The pioneering work AGCN \cite{Li:2018wu} proposes a structure learning model based on distance metric learning.
	Specifically, AGCN computes the generalized Mahalanobis distance between each pair of node features.
Thereafter, it refines the topological structures using a Gaussian kernel of size \(\kappa\) given the distance:
\begin{align}
	\phi(\bm{x}_i, \bm{x}_j) & = \sqrt{(\bm{x}_i - \bm{x}_j)^\top \bm{M} (\bm{x}_i - \bm{x}_j)}, \\
	\widetilde{\bm{A}}_{ij} & = \exp \left(-\frac{\phi(\bm{x}_i, \bm{x}_j)}{2 \kappa^2}\right),
\end{align}
where the symmetric positive semi-definite matrix \(\bm{M} = \bm{W} \bm{W}^\top\) and \(\bm{W} \in \mathbb{R}^{F \times F}\) is a trainable matrix.

\paragraph{Inner-product kernels.}
GRCN \cite{Yu:2020vw}, GAUG-M \cite{Zhao:2021vr}, and CAGNN \cite{Zhu:2020ui} propose to model the edge weights by taking inner product of embeddings of two end nodes:%
\begin{equation}
	\widetilde{\bm{A}} = \sigma(\bm{Z} \bm{Z}^\top),
\end{equation}
where \(\sigma(x) = 1 / (1 + e^{-x})\) is the sigmoid function to normalize the edge weights.
	
\paragraph{Cosine similarity kernels.}
GNN-Guard \cite{Zhang:2020tu} and IDGL \cite{Chen:2020wu} use cosine similarity to measure the edge weights: 
\begin{equation}
	\widetilde{\bm{A}}_{ij} = \cos(\bm{z}_i \odot \bm{w}, \bm{z}_j \odot \bm{w}),
\end{equation}
where \(\bm{w} \in \mathbb{R}^{F'}\) is a trainable parameter.
Recent work HGSL \cite{Zhao:2021vn} extends the idea of IDGL to heterogeneous graphs and jointly considers feature similarity graphs and metapath-induced semantic graphs.

\paragraph{Diffusion kernels.}
GDC \cite{Klicpera:2019vc} employs diffusion kernels to quantify edge connections:
\begin{equation}
	\widetilde{\bm{A}} = \sum_{k=0}^\infty \theta_k \bm{T}^k,
\end{equation}
where \(\theta_k\) is the weighting coefficients satisfying \(\sum_{k=0}^\infty \theta_k = 1\) and \(\bm{T}\) is the generalized transition matrix. %
Popular implementations of \(\bm{T}\) include personalized PageRank \cite{Page:1999wg} and heat kernels \cite{Kondor:2002vc}.
Following GDC, AdaCAD \cite{Lim:2021uu} considers both node features and the graph structure when designing the transition matrix;
GADC \cite{Zhao:2021ad} studies the optimal diffusion step for each node.

\paragraph{Fusion of multiple kernels.}
To enhance the expressivity, AM-GCN \cite{Wang:2020bs} utilizes both cosine and heat kernels to learn informative embeddings in the topology space. %
Similarly, \citet{Yuan:2022sl} employs cosine kernels on node representations and an additional diffusion kernel to encode both local and global aspects of graph structures.

\subsubsection{Neural Approaches}
Compared to metric-based approaches, neural approaches utilize a more complicated deep neural network to model edge weights given node features and representations.
The representative work GLN \cite{Pilco:2019ul} leverages a simple one-layer neural network to iteratively learn graph structures based on local and global node embeddings.
Similarly, GLCN \cite{Jiang:2019wp} implements a single-layer neural network to encode pairwise relationship given two node embeddings.
NeuralSparse \cite{Zheng:2020tp} and PTDNet \cite{Luo:2021wa} leverage multilayer perceptrons to generate an intermediate graph adjacency matrix, followed by discrete sampling.

Apart from simple neural nets, many leverage the attention mechanism to model edge connections due to its ability to capture the complex interaction among nodes.
GAT \cite{Velickovic:2018we} pioneers the use of masked self-attention over one-hop neighborhoods:
\begin{equation}
	\alpha_{ij} = \frac{\exp(\operatorname{LeakyReLU}(\bm{a}^\top [\bm{Wx}_i \| \bm{Wx}_j ]))}{\sum_{k \in \mathcal{N}_i} \exp (\operatorname{LeakyReLU}(\bm{a}^\top [\bm{Wx}_i \| \bm{Wx}_k ]))},
\end{equation}
where \(\bm{W} \in \mathbb{R}^{F' \times F}\) and \(\bm{a} \in \mathbb{R}^{2F'}\) are learnable parameters.
The attentive scores \(\alpha_{ij}\) specify coefficients to different neighboring nodes.
In message propagation, every node computes a weighted combination of the features in its neighborhood:
\begin{equation}
	\bm{z}_i = \operatorname{ReLU}\left(\sum\nolimits_{j \in \mathcal{N}_i} \alpha_{ij} \bm{Wx}_j\right).
\end{equation}
Compared to vanilla GNNs that operate on a static graph structure, the normalized attention scores \(\alpha_{ij}\) dynamically amplifies or attenuates existing connections, which form an adjacency matrix representing the learned graph structures.

One line of improvement of GAT designs different attention mechanisms.
Below we discuss some widely-used work; we refer readers of interest to \citet{Lee:2019do} for a comprehensive survey.
GaAN \cite{Zhang:2018vn} leverages a gated attention mechanism that dynamically adjusts contribution of each attention head.
\citet{Ji:2019hr} propose to use hard and channel-wise attention that improves computational efficiency.
PGN \cite{Velickovic:2020um} and VIB-GSL \cite{Sun:2022wx} propose to use dot-product self-attention to infer dynamic connections.
MAGNA \cite{Wang:2021ci} proposes attention diffusion which incorporates multi-hop context information.
\citet{Brody:2022vy} propose a GAT variant to increase expressiveness of the attention function.

The other line of development generalizes Transformer-like \cite{Vaswani:2017ul} full-attention architectures to the graph domain.
Unlike previous models that only consider local neighborhoods, Transformer performs message propagation between all nodes and encodes the given graph structure as \emph{a soft inductive bias}, which provides more flexibility and is able to discover new relations.
Since message is propagated regardless of graph connectivity, how to restore the positional and structural information (e.g., local connectivity) of the graph becomes a central issue for Transformer-based models.
The earliest work proposes to concatenate Laplacian positional embeddings to node features beforehand \cite{Dwivedi:2020vc}, resembling positional encoding for sequences.
Follow-up work Graphomer \cite{Ying:2021tx} considers additional structural information such as centrality and shortest-path distance.
GraphiT \cite{Mialon:2021ux} proposes relative positional encoding to reweight attention scores by positive definitive kernels and enriches node features with local subgraph structures.
\citet{Kreuzer:2021wi} propose learnable positional encoding, which transforms the Graph's Laplacian spectrum using another Transformer encoder.
Recently, \citet{Dwivedi:2022tc} propose to decouple structural and positional embeddings and enable them to be updated along with other parameters.

\subsubsection{Direct Approaches}
\label{sec:direct-approaches}

Unlike aforementioned approaches, direct approaches treat the adjacency matrix of the target graph as free variables to learn.
Since they do not rely on node representations to model edge connections, direct approaches enjoy more flexibility but have difficulty in learning the matrix parameters.

A large number of direct approaches optimize the adjacency matrix using graph regularizers as mentioned in \S\ref{sec:regularization}, which explicitly specify the property of the optimal graph.
As jointly optimizing adjacency matrix \(\bm{A}^\star\) and model parameters \(\mathbf\Theta\) often involves non-differentiable operators, conventional gradient-based optimization is not applicable.
Concretely, GLNN \cite{Gao:2020em} incorporates the initial graph structure, sparsity and feature smoothness into a hybrid objective.
ProGNN \cite{Jin:2020br} incorporates a low-rank prior which is implemented using the nuclear norm.
Both GLNN and ProGNN use an alternating optimization scheme to iteratively update \(\widetilde{\bm{A}}\) and \(\mathbf\Theta\).
Follow-up work LRGNN \cite{Xu:2021fs} propose an efficient algorithm to speed up the low-rank optimization of adjacency matrix.

Apart from common regularizers, GNNExplainer \cite{Ying:2019wm} introduces an explanation generation loss based on mutual information that identifies the most influential subgraph structure for end tasks:
\begin{equation}
	\min_{\bm{M}} \enspace -\sum_{c=1}^{C} \ind_{y=c} \log P_{\Phi}\left(Y=y \mid \bm{A}_{c} \odot \sigma(\bm{M}), \bm{X}_c\right),
\end{equation}
where \(\ind\) denotes the indicator function, \(\bm{A}_c\) and \(\bm{X}_c\) represent the computation graph and its associated features for label \(c\), and \(\bm{M}\) induces a subgraph for explanation.
Recently, GSML \cite{Wan:2021vo} proposes a meta-learning approach by formulating GSL as a bilevel optimization problem, in which the inner loop represents downstream tasks and the outer loop learns the optimal structure:
\begin{align}
	\quad & \bm{A}^\star = \min_{\bm{A} \in \Phi(\mathcal{G})} \enspace \mathcal{L}_\text{reg}(f_{\mathbf\Theta^\star}(\bm{A}, \bm{X}), \bm{Y}_{U}), \\
	\text { s.t.}\quad & \mathbf\Theta^\star = \argmin_{\mathbf\Theta}  \enspace \mathcal{L}_\text{train}(f_{\mathbf\Theta}(\bm{A}, \bm{X}), \bm{Y}_L),
\end{align}
where \(\Phi(\mathcal{G})\) specifies an admissible topology space that leads to no singleton nodes, and \(\bm{Y}_U\), \(\bm{Y}_L\) are the labels of nodes in the test and training set, respectively. 
Due to lacking supervision of test nodes, GSML proposes solutions to the bilevel optimization via meta-gradients.

The other group of direct approaches model the adjacency matrix in a probabilistic manner, assuming that the intrinsic graph structure is sampled from a certain distribution.
The first work LDS-GNN \cite{Franceschi:2019uz} models the edges between each pair of nodes by sampling from a Bernoulli distribution with learnable parameters and frame GSL as a bilevel programming problem. It utilizes the hypergradient estimation to approximate the solution.
Besides, BGCN \cite{Zhang:2019df} proposes a Bayesian approach which regards the existing graph as a sample from a parametric family of random graphs. Specifically, it utilizes Monte Carlo dropout to sample the learnable parameters in the model for several times on each generated graph.
Similarly, vGCN \cite{Elinas:2020um} also formulates GSL from a Bayesian perspective that considers a prior Bernoulli distribution (parameterized by the observed graph) along with a GNN-based likelihood. Since computation of the posterior distribution is analytically intractable, vGCN utilizes variational inference \cite{Welling:2014tz,Blei:2017el} to approximately infer the posterior.
The posterior parameters are estimated via gradient-based optimization of the Evidence Lower BOund (ELBO), given by
\begin{equation}
	\begin{multlined}
		\mathcal{L}_{\text{ELBO}}(\phi) = \\ \mathbb{E}_{q_{\phi}(\bm{A})} \log p_{\bm{\mathbf\Theta}}(\bm{Y} \mid \bm{X}, \bm{A})-\operatorname{KL}\infdivx{q_{\phi}(\bm{A})}{p(\bm{A})},
	\end{multlined}
\end{equation}
where the first term is the GNN-based likelihood and the second measures the KL divergence between the approximated posterior and the prior.

\subsection{Postprocessing Graph Structures}
\label{sec:postprocessing}

After obtaining an intermediate graph structure \(\widetilde{\bm{A}}\), we consider the following two common postprocessing steps to compute the final adjacency matrix \(\bm{A}^\star\).

\subsubsection{Discrete Sampling}
GSL models involving a sampling step assume the refined graph is generated via an extra sampling process from certain discrete distributions, whose parameters are given by the entries of the intermediate graph adjacency matrix \(\widetilde{\bm{A}}\).
Instead of directly treating this adjacency matrix as edge connection weights in a deterministic manner, the additional sampling step recovers the discrete nature of the graph and gives the structure learner more flexibility to control the property of the final graph, e.g., sparsity.

Note that sampling from a discrete distribution is not differentiable.
Apart from the aforementioned direct approaches (\S\ref{sec:direct-approaches}) that employ special optimization methods, we discuss optimization with conventional gradient descent methods via the reparameterization trick that allows gradients to be backpropagated through the sampling operation \cite{Welling:2014tz,Blei:2017el}.
A common approach is to leverage the Gumbel-Softmax trick \cite{Jang:2017ub} to generate differentiable graphs by drawing samples (i.e. edges) from the Gumbel distribution.
Consider NeuralSparse \cite{Zheng:2020tp} as an example; it approximates the categorical distribution as follows:
\begin{equation}
	\bm{A}^\star_{uv}=\frac{\exp ((\log \widetilde{\bm{A}}_{uv} + \epsilon_{v}) / \tau)}{\sum_{w \in \mathcal{N}_{u}} \exp ((\log \widetilde{\bm{A}}_{uv} + \epsilon_{w}) / \tau)},
\end{equation}
where \(\epsilon_\cdot = -\log (-\log (s))\) is drawn from a Gumbel distribution with noise \(s \sim \operatorname{Uniform}(0, 1)\) and \(\tau\) is a temperature hyperparameter.
To generate a \(k\)-neighbor graph, the above process is repeated without replacement for \(k\) times. %
When \(\tau\) is small, the Gumbel distribution resembles a discrete distribution, which induces strong sparsity on the resulting graph.
GAUG-O \cite{Zhao:2021vr} takes a similar approach with an additional edge prediction loss to stabilize training.

Apart from the Gumbel-Softmax trick, other implementations of end-to-end discrete sampling include the Gumbel-Max trick used by AD-GCL \cite{Suresh:2021vl} and hard concrete sampling \cite{Louizos:2018tl} popularized by PGExplainer \cite{Luo:2020uf} and PTDNet \cite{Luo:2021wa}.
GIB \cite{Wu:2020wy} proposes two model variants that treats the attention weights as the parameters of categorical and Bernoulli distributions, from which samples the refined graph structure.

\subsubsection{Residual Connections}
The initial graph structure, if available, usually carries important prior knowledge on the topology.
It is thus reasonable to assume that the optimized graph structure slightly shifts from the original one.
Based on this assumption, some work adds residual connections \cite{He:2016ib} to incorporate initial states of the graph structure, which is also found to  accelerate and stabilize the training \cite{Li:2018wu}.

Mathematically, the learned edge weights are combined with the original adjacency matrix using an update function:
\begin{equation}
	\bm{A}^{\star} = h(\bm{A}, \widetilde{\bm{A}}).
	\label{eq:residuals}
\end{equation}
In open literature, we may combine the learned adjacency matrix \(\widetilde{\bm{A}}\) with the original structure \(\bm{A}\) as the residuals. A hyperparameter \(\alpha\) is used to mediate the influence of each part, in which Eq. (\ref{eq:residuals}) is instantiated as:
\begin{equation}
	\bm{A}^\star = \alpha\widetilde{\bm{A}} + (1-\alpha)\bm{A}.
\end{equation}
Other popular choices of \(h(\cdot, \cdot)\) include multilayer perceptrons and channel-wise attention networks, which are able to adaptively fuse the two structures.

Note that the original graph structure could be expressed in different forms other than adjacency matrix. For example, AGCN \cite{Li:2018wu} connects the graph Laplacian matrix as the prior knowledge; AM-GCN \cite{Wang:2020bs} propagates node features over the original and learned feature graphs independently and combines both representations to obtain the final node embeddings.

\section{Applications}
\label{sec:applications}

For real-world applications, many graph-based models admit an incomplete/imperfect graph structure and inherently consider the problem of structure learning.
In this section, we briefly review applications of GSL in different domains.

\paragraph{Natural language processing.}
GSL techniques are widely employed to obtain fine-grained linguistic representations in the domain of natural language processing, in which graph structures are constructed by treating words as nodes and connecting them according to both semantic and syntactic patterns.
For information retrieval, \citet{Yu:2021ka} learn hierarchical query-document matching patterns via discarding unimportant words in document graphs.
In relation extraction, \citet{Tian:2021dr} construct a graph based on the linguistic dependency tree and refine the structure by learning different weights for different dependencies.
For sentiment analysis, \citet{Li:2021du} create a semantic graph via computing self-attention among word representations and a syntactic graph based on the dependency tree. Then, they optimize such two graph structures via a differential regularizer, making them capture distinct information.
In question answering, \citet{Yasunaga:2021qa} employ a language-model-based encoder to learn a score for each node so that highlight the most relevant path to the question in the knowledge graph.
For fake news detection, \citet{Xu:2022mf} propose a semantic structure refining scheme to distinguish the beneficial snippets from the redundant ones, thus obtaining the fine-grained evidence for justifying news veracity.

\paragraph{Computer vision and medical imaging.}
In computer vision, wDAE-GNN \cite{Gidaris:2019bf} creates graphs using cosine similarity of the class features to capture the co-dependencies between different classes for few-shot learning; DGCNN \cite{Wang:2019ho} recovers the topology of point-cloud data using GSL, and thus enriches the representation power for both classification and segmentation on point-cloud data.
Another prominent example that applies GSL for imaging data is scene graph generation, which aims to learn relationships between objects. \citet{Qi:2018bn} propose to use convolutions as the structure learner or convolutional LSTMs for spatial-temporal datasets.
Latter approaches propose energy-based objectives to incorporates the structure of scene graphs \cite{Suhail:2021el} or more general constraints during inference \cite{Liu:2022vb}.
For medical image analysis, GPNL \cite{Cosmo:2020fd} leverages metric-based graph structure modeling to learn population graphs representing pairwise patient similarities for disease analysis.
FBNetGen \cite{Kan:2022tg} proposes to learn a functional brain network structure optimized for downstream predictive tasks.

\paragraph{Scientific discovery.}
In scientific discovery domains, e.g., biology, chemistry, graph structures are usually artificial to represent structured interactions within the system. For example, graph structures are constructed for proteins via thresholding the pairwise distance map over all amino acids \cite{Guo:2021gt,Jumper:2021ha}.
In this case, the long-range contacts are usually ignored while building the graphs \cite{Jain:2021rl}.
For optimizing the properties of molecular graphs, we need to learn to weight skeletons of molecules that depict the essential structures of the compounds with optimal properties \cite{Fu:2022ds}.
Similarly, non-bonded interactions are rarely considered while modeling small molecules \cite{Leach:2001mm,Luo:2021pm,Satorras:2021eg}, which may be of great importance to understand the key mechanism of the system.
Graph structure learning allows for learning a more comprehensive representation of the data with minimal information loss and interpretability for scientific discovery.

\section{Challenges and Future Directions}
\label{sec:challenges}

Though promising progress has been made, the development of GSL is still at a nascent stage with many challenges remained unsolved. In this section, we point out some critical problems in current studies and outline several research directions worthy for further investigation.

\paragraph{Beyond homogeneity.} 
Most of the proposed work focuses on homogeneous graphs, where nodes and edges are of the same type. However, nodes and edges in reality are often associated with multiple types \cite{Yang:2021cj}. Few attempts have been made to learn a clean structure on heterogeneous graphs till now. How to learn graph structure by considering complex relations between nodes remains widely open.

\paragraph{Beyond homophily.}
A large body of current work models the edge weights by measuring similarity of node embedding pairs, which is based on the homophily assumption where similar nodes are likely to be connected. However, in real world problems, there are many graphs exhibit strong \emph{heterophily}, where connected nodes are dissimilar, e.g., chemical and molecular graphs \cite{Zhu:2020wt}. There is abundant room for further progress in designing different strategies to learn heterophilous graph structures. 

\paragraph{Learning structures in absence of rich attributes.} 
Most existing work modifies the graph structure based on node embeddings, where rich attributes are indispensable. However, real-world benchmarks carry sparse initial attributes or even do not carry features at all, e.g., in sequential recommendation only user-item interactions are available \cite{Wang:2019iu}. Thereby, the performance of existing methods may be degraded under such scenario. Further work is required to establish the viability of learning a reasonable graph structure when lacking attributes.

\paragraph{Increasing scalability.}
Most of existing work involves the computation of pairwise similarity of node embeddings, which suffers from high computational complexity and thereby hinders the applicability of large-scale datasets \cite{Chen:2020wu}. Further research should be undertaken to increase scalability of GSL methods.

\paragraph{Towards task-agnostic structure learning.}
Existing work most requires task-relevant supervision signal for training. In reality, it is often time-consuming to collect high-quality labels and limited supervision deteriorates the quality of the learned graph structures.
Recently, self-supervised learning on graphs \cite{You:2020ut,Zhu:2021wh} have been developed to tackle such problem and many efforts in supplementing GSL with self-supervision have been made --- SLAPS \cite{Fatemi:2021we} and SUBLIME \cite{Liu:2022uz} to name a few.
However, a principled understanding of the relation between GSL and self-supervised training objectives are still needed.

\section{Concluding Remarks}
\label{sec:conclusion}

In this paper, we have broadly reviewed existing studies of Graph Structure Learning (GSL).
We first elaborate the concept of GSL and formulate a general pipeline.
Then, we categorize recent work into three groups: metric-based approaches, neural approaches, and direct approaches.
We summarize key characteristics in each type and discuss common structure modeling techniques.
Furthermore, we discuss applications of GSL in different domains.
Finally, we outline challenges in current studies and point out directions for further research.
We envision this survey to help expand our understanding of GSL and to guide future development of GSL algorithms.

\bibliographystyle{named}
\bibliography{ijcai22}

\begin{thebibliography}{}

\bibitem[\protect\citeauthoryear{Beck and Teboulle}{2010}]{Beck:2010ga}
A. Beck and M. Teboulle.
\newblock {Gradient-Based Algorithms with Applications to Signal Recovery
  Problems}.
\newblock In {\em Convex Optimization in Signal Processing and Communications}.
  2010.

\bibitem[\protect\citeauthoryear{Bentley \bgroup \em et~al.\@\egroup
  }{1977}]{Bentley:1977kt}
J.~L. Bentley, D.~F. Stanat, and E.~H. Williams~Jr.
\newblock {The Complexity of Finding Fixed-Radius Near Neighbors}.
\newblock {\em Inf. Process. Lett.}, 1977.

\bibitem[\protect\citeauthoryear{Biggs \bgroup \em et~al.\@\egroup
  }{1993}]{Biggs:1993ag}
N. Biggs, N.~L. Biggs, and B. Norman.
\newblock {\em {Algebraic Graph Theory}}.
\newblock Cambridge University Press, 1993.

\bibitem[\protect\citeauthoryear{Blei \bgroup \em et~al.\@\egroup
  }{2017}]{Blei:2017el}
D.~M. Blei, A. Kucukelbir, and J.~D. McAuliffe.
\newblock {Variational Inference: A Review for Statisticians}.
\newblock {\em J. Am. Stat. Assoc.}, 2017.

\bibitem[\protect\citeauthoryear{Brody \bgroup \em et~al.\@\egroup
  }{2022}]{Brody:2022vy}
S. Brody, U. Alon, and E. Yahav.
\newblock {How Attentive are Graph Attention Networks?}
\newblock 2022.

\bibitem[\protect\citeauthoryear{Cai \bgroup \em et~al.\@\egroup
  }{2010}]{Cai:2010hz}
J.-F. Cai, E.~J. Cand{\`e}s, and Z. Shen.
\newblock {A Singular Value Thresholding Algorithm for Matrix Completion}.
\newblock {\em SIAM J. Optim.}, 2010.

\bibitem[\protect\citeauthoryear{Chen \bgroup \em et~al.\@\egroup
  }{2020}]{Chen:2020wu}
Y. Chen, L. Wu, and M.~J. Zaki.
\newblock {Iterative Deep Graph Learning for Graph Neural Networks: Better and
  Robust Node Embeddings}.
\newblock In {\em NeurIPS}, 2020.

\bibitem[\protect\citeauthoryear{Chung}{1997}]{Chung:1997sp}
F.~R. Chung.
\newblock {\em {Spectral Graph Theory}}.
\newblock AMS, 1997.

\bibitem[\protect\citeauthoryear{Cosmo \bgroup \em et~al.\@\egroup
  }{2020}]{Cosmo:2020fd}
L. Cosmo, A. Kazi, S.-A. Ahmadi, N. Navab, and M. Bronstein.
\newblock {Latent Patient Network Learning for Automatic Diagnosis}.
\newblock In {\em MICCAI}, 2020.

\bibitem[\protect\citeauthoryear{Dai \bgroup \em et~al.\@\egroup
  }{2018}]{Dai:2018ts}
H. Dai, H. Li, T. Tian, X. Huang, L. Wang, J. Zhu, and L. Song.
\newblock {Adversarial Attack on Graph Structured Data}.
\newblock In {\em ICML}, 2018.

\bibitem[\protect\citeauthoryear{Dong \bgroup \em et~al.\@\egroup
  }{2019}]{Dong:2019gq}
X. Dong, D. Thanou, M.~G. Rabbat, and P. Frossard.
\newblock {Learning Graphs From Data: A Signal Representation Perspective}.
\newblock {\em IEEE Signal Process. Mag.}, 2019.

\bibitem[\protect\citeauthoryear{Du \bgroup \em et~al.\@\egroup
  }{2021}]{Du:2021gt}
Y. Du, S. Wang, X. Guo, H. Cao, S. Hu, J. Jiang, A. Varala, A. Angirekula, and
  L. Zhao.
\newblock {GraphGT: Machine Learning Datasets for Deep Graph Generation and
  Transformation}.
\newblock In {\em NeurIPS Datasets and Benchmarks}, 2021.

\bibitem[\protect\citeauthoryear{Dwivedi and Bresson}{2020}]{Dwivedi:2020vc}
V.~P. Dwivedi and X. Bresson.
\newblock {A Generalization of Transformer Networks to Graphs}.
\newblock In {\em DLG@AAAI}, 2020.

\bibitem[\protect\citeauthoryear{Dwivedi \bgroup \em et~al.\@\egroup
  }{2022}]{Dwivedi:2022tc}
V.~P. Dwivedi, A.~T. Luu, T. Laurent, Y. Bengio, and X. Bresson.
\newblock {Graph Neural Networks with Learnable Structural and Positional
  Representations}.
\newblock 2022.

\bibitem[\protect\citeauthoryear{Elinas \bgroup \em et~al.\@\egroup
  }{2020}]{Elinas:2020um}
P. Elinas, E.~V. Bonilla, and L.~C. Tiao.
\newblock {Variational Inference for Graph Convolutional Networks in the
  Absence of Graph Data and Adversarial Settings}.
\newblock In {\em NeurIPS}, 2020.

\bibitem[\protect\citeauthoryear{Fatemi \bgroup \em et~al.\@\egroup
  }{2021}]{Fatemi:2021we}
B. Fatemi, L. El~Asri, and S.~M. Kazemi.
\newblock {SLAPS: Self-Supervision Improves Structure Learning for Graph Neural
  Networks}.
\newblock In {\em NeurIPS}, 2021.

\bibitem[\protect\citeauthoryear{Franceschi \bgroup \em et~al.\@\egroup
  }{2019}]{Franceschi:2019uz}
L. Franceschi, M. Niepert, M. Pontil, and X. He.
\newblock {Learning Discrete Structures for Graph Neural Networks}.
\newblock In {\em ICML}, 2019.

\bibitem[\protect\citeauthoryear{Fu \bgroup \em et~al.\@\egroup
  }{2022}]{Fu:2022ds}
T. Fu, W. Gao, C. Xiao, J. Yasonik, C.~W. Coley, and J. Sun.
\newblock {Differentiable Scaffolding Tree for Molecule Optimization}.
\newblock In {\em ICLR}, 2022.

\bibitem[\protect\citeauthoryear{Fuel \bgroup \em et~al.\@\egroup
  }{2001}]{Fazel:2001rm}
M. Fuel, H. Hindi, and S.~P. Boyd.
\newblock {A Rank Minimization Heuristic with Application to Minimum Order
  System Approximation}.
\newblock In {\em ACC}, 2001.

\bibitem[\protect\citeauthoryear{Gao and Ji}{2019}]{Ji:2019hr}
H. Gao and S. Ji.
\newblock {Graph Representation Learning via Hard and Channel-Wise Attention
  Networks}.
\newblock In {\em KDD}, 2019.

\bibitem[\protect\citeauthoryear{Gao \bgroup \em et~al.\@\egroup
  }{2020}]{Gao:2020em}
X. Gao, W. Hu, and Z. Guo.
\newblock {Exploring Structure-Adaptive Graph Learning for Robust
  Semi-Supervised Classification}.
\newblock In {\em ICME}, 2020.

\bibitem[\protect\citeauthoryear{Gidaris and Komodakis}{2019}]{Gidaris:2019bf}
S. Gidaris and N. Komodakis.
\newblock {Generating Classification Weights With GNN Denoising Autoencoders
  for Few-Shot Learning}.
\newblock In {\em CVPR}, 2019.

\bibitem[\protect\citeauthoryear{Gilmer \bgroup \em et~al.\@\egroup
  }{2017}]{Gilmer:2017tl}
J. Gilmer, S.~S. Schoenholz, P.~F. Riley, O. Vinyals, and G.~E. Dahl.
\newblock {Neural Message Passing for Quantum Chemistry}.
\newblock In {\em ICML}, 2017.

\bibitem[\protect\citeauthoryear{Guo \bgroup \em et~al.\@\egroup
  }{2021}]{Guo:2021gt}
X. Guo, Y. Du, S. Tadepalli, L. Zhao, and A. Shehu.
\newblock {Generating Tertiary Protein Structures via Interpretable Graph
  Variational Autoencoders}.
\newblock {\em Bioinfo. Adv.}, 2021.

\bibitem[\protect\citeauthoryear{He \bgroup \em et~al.\@\egroup
  }{2016}]{He:2016ib}
K. He, X. Zhang, S. Ren, and J. Sun.
\newblock {Deep Residual Learning for Image Recognition}.
\newblock In {\em CVPR}, 2016.

\bibitem[\protect\citeauthoryear{Jain \bgroup \em et~al.\@\egroup
  }{2021}]{Jain:2021rl}
P. Jain, Z. Wu, M. Wright, A. Mirhoseini, J.~E. Gonzalez, and I. Stoica.
\newblock {Representing Long-Range Context for Graph Neural Networks with
  Global Attention}.
\newblock In {\em NeurIPS}, 2021.

\bibitem[\protect\citeauthoryear{Jang \bgroup \em et~al.\@\egroup
  }{2017}]{Jang:2017ub}
E. Jang, S. Gu, and B. Poole.
\newblock {Categorical Reparameterization with Gumbel-Softmax}.
\newblock In {\em ICLR}, 2017.

\bibitem[\protect\citeauthoryear{Jiang \bgroup \em et~al.\@\egroup
  }{2019}]{Jiang:2019wp}
B. Jiang, Z. Zhang, D. Lin, and J. Tang.
\newblock {Semi-supervised Learning with Graph Learning-Convolutional
  Networks}.
\newblock In {\em CVPR}, 2019.

\bibitem[\protect\citeauthoryear{Jin \bgroup \em et~al.\@\egroup
  }{2020}]{Jin:2020br}
W. Jin, Y. Ma, X. Liu, X. Tang, S. Wang, and J. Tang.
\newblock {Graph Structure Learning for Robust Graph Neural Networks}.
\newblock In {\em KDD}, 2020.

\bibitem[\protect\citeauthoryear{Jumper \bgroup \em et~al.\@\egroup
  }{2021}]{Jumper:2021ha}
J. Jumper, R. Evans, A. Pritzel, T. Green, M. Figurnov, O. Ronneberger, K.
  Tunyasuvunakool, R. Bates, A. {\v{Z}}{\'\i}dek, A. Potapenko, et~al.
\newblock {Highly Accurate Protein Structure Prediction with AlphaFold}.
\newblock {\em Nature}, 2021.

\bibitem[\protect\citeauthoryear{Kalofolias}{2016}]{Kalofolias:2016tf}
V. Kalofolias.
\newblock {How to Learn a Graph from Smooth Signals}.
\newblock In {\em AISTATS}, 2016.

\bibitem[\protect\citeauthoryear{Kan \bgroup \em et~al.\@\egroup
  }{2022}]{Kan:2022tg}
X. Kan, H. Cui, J. Lukemire, Y. Guo, and C. Yang.
\newblock {FBNetGen: Task-aware GNN-based fMRI Analysis via Functional Brain
  Network Generation}.
\newblock In {\em MIDL}, 2022.

\bibitem[\protect\citeauthoryear{Kingma and Welling}{2014}]{Welling:2014tz}
D.~P. Kingma and M. Welling.
\newblock {Auto-Encoding Variational Bayes}.
\newblock In {\em ICLR}, 2014.

\bibitem[\protect\citeauthoryear{Klicpera \bgroup \em et~al.\@\egroup
  }{2019}]{Klicpera:2019vc}
J. Klicpera, S. Wei{\ss}enberger, and S. G{\"u}nnemann.
\newblock {Diffusion Improves Graph Learning}.
\newblock In {\em NeurIPS}, 2019.

\bibitem[\protect\citeauthoryear{Kondor and Lafferty}{2002}]{Kondor:2002vc}
R. Kondor and J.~D. Lafferty.
\newblock {Diffusion Kernels on Graphs and Other Discrete Input Spaces}.
\newblock In {\em ICLR}, 2002.

\bibitem[\protect\citeauthoryear{Kreuzer \bgroup \em et~al.\@\egroup
  }{2021}]{Kreuzer:2021wi}
D. Kreuzer, D. Beaini, W.~L. Hamilton, V. L{\'e}tourneau, and P. Tossou.
\newblock {Rethinking Graph Transformers with Spectral Attention}.
\newblock In {\em NeurIPS}, 2021.

\bibitem[\protect\citeauthoryear{Leach}{2001}]{Leach:2001mm}
A.~R. Leach.
\newblock {\em {Molecular Modelling: Principles and Applications}}.
\newblock Pearson Education, 2001.

\bibitem[\protect\citeauthoryear{Lee \bgroup \em et~al.\@\egroup
  }{2019}]{Lee:2019do}
J.~B. Lee, R.~A. Rossi, S. Kim, N.~K. Ahmed, and E. Koh.
\newblock {Attention Models in Graphs: A Survey}.
\newblock {\em TKDD}, 2019.

\bibitem[\protect\citeauthoryear{Li \bgroup \em et~al.\@\egroup
  }{2018}]{Li:2018wu}
R. Li, S. Wang, F. Zhu, and J. Huang.
\newblock {Adaptive Graph Convolutional Neural Networks}.
\newblock In {\em AAAI}, 2018.

\bibitem[\protect\citeauthoryear{Li \bgroup \em et~al.\@\egroup
  }{2021a}]{Li:2021du}
R. Li, H. Chen, F. Feng, Z. Ma, X. Wang, and E. Hovy.
\newblock {Dual Graph Convolutional Networks for Aspect-based Sentiment
  Analysis}.
\newblock In {\em ACL}, 2021.

\bibitem[\protect\citeauthoryear{Li \bgroup \em et~al.\@\egroup
  }{2021b}]{Li:2021fa}
X. Li, Y. Zhou, N.~C. Dvornek, M. Zhang, S. Gao, J. Zhuang, D. Scheinost, L.~H.
  Staib, P. Ventola, and J.~S. Duncan.
\newblock {BrainGNN: Interpretable Brain Graph Neural Network for fMRI
  Analysis}.
\newblock {\em Med. Image Anal.}, 2021.

\bibitem[\protect\citeauthoryear{Lim \bgroup \em et~al.\@\egroup
  }{2021}]{Lim:2021uu}
J. Lim, D. Um, H.~J. Chang, D.~U. Jo, and J.~Y. Choi.
\newblock {Class-Attentive Diffusion Network for Semi-Supervised
  Classification}.
\newblock In {\em AAAI}, 2021.

\bibitem[\protect\citeauthoryear{Liu \bgroup \em et~al.\@\egroup
  }{2022a}]{Liu:2022vb}
D. Liu, M. Bober, and J. Kittler.
\newblock {Constrained Structure Learning for Scene Graph Generation}.
\newblock {\em arXiv.org}, January 2022.

\bibitem[\protect\citeauthoryear{Liu \bgroup \em et~al.\@\egroup
  }{2022b}]{Liu:2022uz}
Y. Liu, Y. Zheng, D. Zhang, H. Chen, H. Peng, and S. Pan.
\newblock {Towards Unsupervised Deep Graph Structure Learning}.
\newblock In {\em WWW}, 2022.

\bibitem[\protect\citeauthoryear{Louizos \bgroup \em et~al.\@\egroup
  }{2018}]{Louizos:2018tl}
C. Louizos, M. Welling, and D.~P. Kingma.
\newblock {Learning Sparse Neural Networks through L0 Regularization}.
\newblock In {\em ICLR}, 2018.

\bibitem[\protect\citeauthoryear{Luo \bgroup \em et~al.\@\egroup
  }{2020}]{Luo:2020uf}
D. Luo, W. Cheng, D. Xu, W. Yu, B. Zong, H. Chen, and X. Zhang.
\newblock {Parameterized Explainer for Graph Neural Network}.
\newblock In {\em NeurIPS}, 2020.

\bibitem[\protect\citeauthoryear{Luo \bgroup \em et~al.\@\egroup
  }{2021a}]{Luo:2021wa}
D. Luo, W. Cheng, W. Yu, B. Zong, J. Ni, H. Chen, and X. Zhang.
\newblock {Learning to Drop: Robust Graph Neural Network via Topological
  Denoising}.
\newblock In {\em WSDM}, 2021.

\bibitem[\protect\citeauthoryear{Luo \bgroup \em et~al.\@\egroup
  }{2021b}]{Luo:2021pm}
S. Luo, C. Shi, M. Xu, and J. Tang.
\newblock {Predicting Molecular Conformation via Dynamic Graph Score Matching}.
\newblock In {\em NeurIPS}, 2021.

\bibitem[\protect\citeauthoryear{Mialon \bgroup \em et~al.\@\egroup
  }{2021}]{Mialon:2021ux}
G. Mialon, D. Chen, M. Selosse, and J. Mairal.
\newblock {GraphiT: Encoding Graph Structure in Transformers}.
\newblock {\em arXiv.org}, June 2021.

\bibitem[\protect\citeauthoryear{Newman}{2018}]{Newman:2018ne}
M. Newman.
\newblock {\em {Networks (Second Edition)}}.
\newblock Oxford University Press, 2018.

\bibitem[\protect\citeauthoryear{Ortega \bgroup \em et~al.\@\egroup
  }{2018}]{Ortega:2018ea}
A. Ortega, P. Frossard, J. Kovacevic, J.~M.~F. Moura, and P. Vandergheynst.
\newblock {Graph Signal Processing: Overview, Challenges, and Applications}.
\newblock {\em Proc. IEEE}, 2018.

\bibitem[\protect\citeauthoryear{Page \bgroup \em et~al.\@\egroup
  }{1999}]{Page:1999wg}
L. Page, S. Brin, R. Motwani, and T. Winograd.
\newblock {The PageRank Citation Ranking: Bringing Order to the Web}.
\newblock Technical report, November 1999.

\bibitem[\protect\citeauthoryear{Pilco and Rivera}{2019}]{Pilco:2019ul}
D.~S. Pilco and A.~R. Rivera.
\newblock {Graph Learning Network: A Structure Learning Algorithm}.
\newblock In {\em LRGD@ICML}, 2019.

\bibitem[\protect\citeauthoryear{Preparata and Shamos}{1985}]{Preparata:1985cg}
F.~P. Preparata and M.~I. Shamos.
\newblock {\em {Computational Geometry: An Introduction}}.
\newblock Springer, 1985.

\bibitem[\protect\citeauthoryear{Qi \bgroup \em et~al.\@\egroup
  }{2018}]{Qi:2018bn}
S. Qi, W. Wang, B. Jia, J. Shen, and S.-C. Zhu.
\newblock {Learning Human-Object Interactions by Graph Parsing Neural
  Networks}.
\newblock In {\em ECCV}, 2018.

\bibitem[\protect\citeauthoryear{Satorras \bgroup \em et~al.\@\egroup
  }{2021}]{Satorras:2021eg}
V.~G. Satorras, E. Hoogeboom, and M. Welling.
\newblock E(n) equivariant graph neural networks.
\newblock In {\em ICML}, 2021.

\bibitem[\protect\citeauthoryear{Suhail \bgroup \em et~al.\@\egroup
  }{2021}]{Suhail:2021el}
M. Suhail, A. Mittal, B. Siddiquie, C. Broaddus, J. Eledath, G. Medioni, and L.
  Sigal.
\newblock {Energy-based Learning for Scene Graph Generation}.
\newblock In {\em CVPR}, 2021.

\bibitem[\protect\citeauthoryear{Sun \bgroup \em et~al.\@\egroup
  }{2022}]{Sun:2022wx}
Q. Sun, J. Li, H. Peng, J. Wu, X. Fu, C. Ji, and P.~S. Yu.
\newblock {Graph Structure Learning with Variational Information Bottleneck}.
\newblock In {\em AAAI}, 2022.

\bibitem[\protect\citeauthoryear{Suresh \bgroup \em et~al.\@\egroup
  }{2021}]{Suresh:2021vl}
S. Suresh, P. Li, C. Hao, and J. Neville.
\newblock {Adversarial Graph Augmentation to Improve Graph Contrastive
  Learning}.
\newblock In {\em NeurIPS}, 2021.

\bibitem[\protect\citeauthoryear{Tian \bgroup \em et~al.\@\egroup
  }{2021}]{Tian:2021dr}
Y. Tian, G. Chen, Y. Song, and X. Wan.
\newblock {Dependency-driven Relation Extraction with Attentive Graph
  Convolutional Networks}.
\newblock In {\em ACL}, 2021.

\bibitem[\protect\citeauthoryear{Vaswani \bgroup \em et~al.\@\egroup
  }{2017}]{Vaswani:2017ul}
A. Vaswani, N. Shazeer, N. Parmar, J. Uszkoreit, L. Jones, A.~N. Gomez, U.
  Kaiser, and I. Polosukhin.
\newblock {Attention is All You Need}.
\newblock In {\em NIPS}, 2017.

\bibitem[\protect\citeauthoryear{Veli{\v c}kovi{\'c} \bgroup \em
  et~al.\@\egroup }{2018}]{Velickovic:2018we}
P. Veli{\v c}kovi{\'c}, G. Cucurull, A. Casanova, A. Romero, P. Li{\`o}, and Y.
  Bengio.
\newblock {Graph Attention Networks}.
\newblock In {\em ICLR}, 2018.

\bibitem[\protect\citeauthoryear{Veli{\v c}kovi{\'c} \bgroup \em
  et~al.\@\egroup }{2020}]{Velickovic:2020um}
P. Veli{\v c}kovi{\'c}, L. Buesing, M.~C. Overlan, R. Pascanu, O. Vinyals, and
  C. Blundell.
\newblock {Pointer Graph Networks}.
\newblock In {\em NeurIPS}, 2020.

\bibitem[\protect\citeauthoryear{Wan and Kokel}{2021}]{Wan:2021vo}
G. Wan and H. Kokel.
\newblock {Graph Sparsification via Meta-Learning}.
\newblock In {\em DLG@AAAI}, 2021.

\bibitem[\protect\citeauthoryear{Wang \bgroup \em et~al.\@\egroup
  }{2019a}]{Wang:2019iu}
S. Wang, L. Hu, Y. Wang, L. Cao, Q.~Z. Sheng, and M.~A. Orgun.
\newblock {Sequential Recommender Systems: Challenges, Progress and Prospects}.
\newblock In {\em IJCAI}, 2019.

\bibitem[\protect\citeauthoryear{Wang \bgroup \em et~al.\@\egroup
  }{2019b}]{Wang:2019ho}
Y. Wang, Y. Sun, Z. Liu, S.~E. Sarma, M.~M. Bronstein, and J.~M. Solomon.
\newblock {Dynamic Graph CNN for Learning on Point Clouds}.
\newblock {\em TOG}, 2019.

\bibitem[\protect\citeauthoryear{Wang \bgroup \em et~al.\@\egroup
  }{2020}]{Wang:2020bs}
X. Wang, M. Zhu, D. Bo, P. Cui, C. Shi, and J. Pei.
\newblock {AM-GCN: Adaptive Multi-channel Graph Convolutional Networks}.
\newblock In {\em KDD}, 2020.

\bibitem[\protect\citeauthoryear{Wang \bgroup \em et~al.\@\egroup
  }{2021}]{Wang:2021ci}
G. Wang, R. Ying, J. Huang, and J. Leskovec.
\newblock {Multi-hop Attention Graph Neural Networks}.
\newblock In {\em IJCAI}, 2021.

\bibitem[\protect\citeauthoryear{Wu and Lange}{2008}]{Wu:2008cd}
T.~T. Wu and K. Lange.
\newblock {Coordinate Descent Algorithms for Lasso Penalized Regression}.
\newblock {\em Ann. Appl. Stat.}, 2008.

\bibitem[\protect\citeauthoryear{Wu \bgroup \em et~al.\@\egroup
  }{2020}]{Wu:2020wy}
T. Wu, H. Ren, P. Li, and J. Leskovec.
\newblock {Graph Information Bottleneck}.
\newblock In {\em NeurIPS}, 2020.

\bibitem[\protect\citeauthoryear{Xu \bgroup \em et~al.\@\egroup
  }{2021}]{Xu:2021fs}
H. Xu, L. Xiang, J. Yu, A. Cao, and X. Wang.
\newblock {Speedup Robust Graph Structure Learning with Low-Rank Information}.
\newblock In {\em CIKM}, 2021.

\bibitem[\protect\citeauthoryear{Xu \bgroup \em et~al.\@\egroup
  }{2022}]{Xu:2022mf}
W. Xu, J. Wu, Q. Liu, S. Wu, and L. Wang.
\newblock {Mining Fine-grained Semantics via Graph Neural Networks for
  Evidence-based Fake News Detection}.
\newblock In {\em WWW}, 2022.

\bibitem[\protect\citeauthoryear{Yang \bgroup \em et~al.\@\egroup
  }{2021}]{Yang:2021cj}
C. Yang, Y. Xiao, Y. Zhang, Y. Sun, and J. Han.
\newblock {Heterogeneous Network Representation Learning: A Unified Framework
  with Survey and Benchmark}.
\newblock {\em TKDE}, 2021.

\bibitem[\protect\citeauthoryear{Yasunaga \bgroup \em et~al.\@\egroup
  }{2021}]{Yasunaga:2021qa}
M. Yasunaga, H. Ren, A. Bosselut, P. Liang, and J. Leskovec.
\newblock {QA-GNN: Reasoning with Language Models and Knowledge Graphs for
  Question Answering}.
\newblock In {\em NAACL}, 2021.

\bibitem[\protect\citeauthoryear{Ying \bgroup \em et~al.\@\egroup
  }{2019}]{Ying:2019wm}
R. Ying, D. Bourgeois, J. You, M. Zitnik, and J. Leskovec.
\newblock {GNN Explainer: A Tool for Post-hoc Explanation of Graph Neural
  Networks}.
\newblock In {\em NeurIPS}, 2019.

\bibitem[\protect\citeauthoryear{Ying \bgroup \em et~al.\@\egroup
  }{2021}]{Ying:2021tx}
C. Ying, T. Cai, S. Luo, S. Zheng, G. Ke, D. He, Y. Shen, and T.-Y. Liu.
\newblock {Do Transformers Really Perform Bad for Graph Representation?}
\newblock In {\em NeurIPS}, 2021.

\bibitem[\protect\citeauthoryear{You \bgroup \em et~al.\@\egroup
  }{2020}]{You:2020ut}
Y. You, T. Chen, Y. Sui, T. Chen, Z. Wang, and Y. Shen.
\newblock {Graph Contrastive Learning with Augmentations}.
\newblock In {\em NeurIPS}, 2020.

\bibitem[\protect\citeauthoryear{Yu \bgroup \em et~al.\@\egroup
  }{2020}]{Yu:2020vw}
D. Yu, R. Zhang, Z. Jiang, Y. Wu, and Y. Yang.
\newblock {Graph-Revised Convolutional Network}.
\newblock In {\em ECML PKDD}, 2020.

\bibitem[\protect\citeauthoryear{Yu \bgroup \em et~al.\@\egroup
  }{2021}]{Yu:2021ka}
X. Yu, W. Xu, Z. Cui, S. Wu, and L. Wang.
\newblock {Graph-based Hierarchical Relevance Matching Signals for Ad-hoc
  Retrieval}.
\newblock In {\em WWW}, 2021.

\bibitem[\protect\citeauthoryear{Yuan \bgroup \em et~al.\@\egroup
  }{2022}]{Yuan:2022sl}
J. Yuan, M. Cao, H. Cheng, H. Yu, J. Xie, and C. Wang.
\newblock {A Unified Structure Learning Framework for Graph Attention
  Networks}.
\newblock {\em Neurocomputing}, 2022.

\bibitem[\protect\citeauthoryear{Zhang and Zitnik}{2020}]{Zhang:2020tu}
X. Zhang and M. Zitnik.
\newblock {GNNGuard: Defending Graph Neural Networks against Adversarial
  Attacks}.
\newblock In {\em NeurIPS}, 2020.

\bibitem[\protect\citeauthoryear{Zhang \bgroup \em et~al.\@\egroup
  }{2018}]{Zhang:2018vn}
J. Zhang, X. Shi, J. Xie, H. Ma, I. King, and D.-Y. Yeung.
\newblock {GaAN: Gated Attention Networks for Learning on Large and
  Spatiotemporal Graphs}.
\newblock In {\em UAI}, 2018.

\bibitem[\protect\citeauthoryear{Zhang \bgroup \em et~al.\@\egroup
  }{2019}]{Zhang:2019df}
Y. Zhang, S. Pal, M. Coates, and D. {\"U}stebay.
\newblock {Bayesian Graph Convolutional Neural Networks for Semi-Supervised
  Classification}.
\newblock In {\em AAAI}, 2019.

\bibitem[\protect\citeauthoryear{Zhao \bgroup \em et~al.\@\egroup
  }{2021a}]{Zhao:2021ad}
J. Zhao, Y. Dong, M. Ding, E. Kharlamov, and J. Tang.
\newblock {Adaptive Diffusion in Graph Neural Networks}.
\newblock In {\em NeurIPS}, 2021.

\bibitem[\protect\citeauthoryear{Zhao \bgroup \em et~al.\@\egroup
  }{2021b}]{Zhao:2021vn}
J. Zhao, X. Wang, C. Shi, B. Hu, G. Song, and Y. Ye.
\newblock {Heterogeneous Graph Structure Learning for Graph Neural Networks}.
\newblock In {\em AAAI}, 2021.

\bibitem[\protect\citeauthoryear{Zhao \bgroup \em et~al.\@\egroup
  }{2021c}]{Zhao:2021vr}
T. Zhao, Y. Liu, L. Neves, O. Woodford, M. Jiang, and N. Shah.
\newblock {Data Augmentation for Graph Neural Networks}.
\newblock In {\em AAAI}, 2021.

\bibitem[\protect\citeauthoryear{Zheng \bgroup \em et~al.\@\egroup
  }{2020}]{Zheng:2020tp}
C. Zheng, B. Zong, W. Cheng, D. Song, J. Ni, W. Yu, H. Chen, and W. Wang.
\newblock {Robust Graph Representation Learning via Neural Sparsification}.
\newblock In {\em ICML}, 2020.

\bibitem[\protect\citeauthoryear{Zhu \bgroup \em et~al.\@\egroup
  }{2019}]{Zhu:2019ik}
D. Zhu, Z. Zhang, P. Cui, and W. Zhu.
\newblock {Robust Graph Convolutional Networks Against Adversarial Attacks}.
\newblock In {\em KDD}, 2019.

\bibitem[\protect\citeauthoryear{Zhu \bgroup \em et~al.\@\egroup
  }{2020a}]{Zhu:2020wt}
J. Zhu, Y. Yan, L. Zhao, M. Heimann, L. Akoglu, and D. Koutra.
\newblock {Beyond Homophily in Graph Neural Networks: Current Limitations and
  Effective Designs}.
\newblock In {\em NeurIPS}, 2020.

\bibitem[\protect\citeauthoryear{Zhu \bgroup \em et~al.\@\egroup
  }{2020b}]{Zhu:2020ui}
Y. Zhu, Y. Xu, F. Yu, S. Wu, and L. Wang.
\newblock {CAGNN: Cluster-Aware Graph Neural Networks for Unsupervised Graph
  Representation Learning}.
\newblock {\em arXiv.org}, September 2020.

\bibitem[\protect\citeauthoryear{Zhu \bgroup \em et~al.\@\egroup
  }{2021}]{Zhu:2021wh}
Y. Zhu, Y. Xu, F. Yu, Q. Liu, S. Wu, and L. Wang.
\newblock {Graph Contrastive Learning with Adaptive Augmentation}.
\newblock In {\em WWW}, 2021.

\end{thebibliography}

\end{document}